\documentclass[letterpaper, 10 pt, conference]{ieeeconf}

\IEEEoverridecommandlockouts 
\usepackage{graphicx}
\usepackage{epsfig} 
\usepackage{amsmath} 

\usepackage{amssymb}  
\usepackage{amsthm}
\usepackage{wasysym}
\usepackage{mathrsfs}
\usepackage{bbm}
\usepackage{color}
\usepackage[ruled,vlined,linesnumbered]{algorithm2e}
\usepackage[colorlinks=true,linkcolor=black,anchorcolor=black,citecolor=black,filecolor=black,menucolor=black,runcolor=black,urlcolor=black]{hyperref}
\usepackage{etoolbox}
\usepackage[table,xcdraw,dvipsnames]{xcolor}
\usepackage{multirow}
\usepackage{mathtools}
\usepackage{subfig}
\usepackage{caption}
\usepackage{bm}
\usepackage{array}
\usepackage{booktabs}
\newcolumntype{?}{!{\vrule width 2pt}}

\usepackage{siunitx} 
\usepackage[capitalize,noabbrev,nameinlink]{cleveref}
\crefformat{equation}{(#2#1#3)}
\usepackage[acronyms, shortcuts]{glossaries}

\usepackage{outlines}
\let\labelindent\relax
\usepackage{enumitem}
\setenumerate[1]{label=\Roman*.}
\setenumerate[2]{label=\Alph*.}
\setenumerate[3]{label=\arabic*.}
\setenumerate[4]{label=\alph*.}

\usepackage{algorithm2e}
\RestyleAlgo{ruled}

\glsdisablehyper

\newacronym{ssta}{SSTA}{self-supervised traffic advisor}
\newacronym{cav}{CAV}{connected automated vehicle}

\newtheorem{defn}{Definition}
\newtheorem{rem}[defn]{Remark}

\providecommand{\R}{\ensuremath \mathbb{R}}

\providecommand{\N}{\ensuremath \mathbb{N}}

\newcommand{\regtext}[1]{\mathrm{\textnormal{#1}}}

\newcommand{\ts}[1]{\textsuperscript{#1}}
\newcommand{\code}[1]{{\texttt{#1}}}
\newcommand{\ith}{$i$\ts{th}\xspace}

\newcommand{\norm}[1]{\left\Vert#1\right\Vert}

\newcommand{\davg}{\bar{d}}
\newcommand{\tfirst}{t^*}

\newcommand{\zeros}{\mathbf{0}}

\newcommand{\lbl}[1]{_{\regtext{#1}}}
\newcommand{\goal}{\lbl{goal}}

\newcommand{\loss}{\mathcal{L}}

\usepackage[
    style=ieee,
    doi=false,
    isbn=false,
    url=false,
    eprint=false,
    backend=biber,
    natbib=true
    ]{biblatex}

\bibliography{references}

\title{\LARGE \bf
Connected Autonomous Vehicle Motion Planning with Video Predictions from Smart, Self-Supervised Infrastructure
}

\author{
Jiankai Sun\ts{1},
Shreyas Kousik\ts{2},
David Fridovich-Keil\ts{3}, and
Mac Schwager\ts{1}
\thanks{\ts{1}Stanford University,
\ts{2}Georgia Institute of Technology,
\ts{3}University of Texas at Austin.
}
}

\begin{document}
\maketitle
\thispagestyle{plain}
\pagestyle{plain}

\begin{abstract}
Connected autonomous vehicles (CAVs) promise to enhance safety, efficiency, and sustainability in urban transportation.
However, this is contingent upon a CAV correctly predicting the motion of surrounding agents and planning its own motion safely.
Doing so is challenging in complex urban environments due to frequent occlusions and interactions among many agents.
One solution is to leverage smart infrastructure to augment a CAV's situational awareness; the present work leverages a recently-proposed ``Self-Supervised Traffic Advisor'' (SSTA) framework of smart sensors that teach themselves to generate and broadcast useful video predictions of road users.
In this work, SSTA predictions are modified to predict future occupancy instead of raw video, which reduces the data footprint of broadcast predictions.
The resulting predictions are used within a planning framework, demonstrating that this design can effectively aid CAV motion planning.
A variety of numerical experiments study the key factors that make SSTA outputs useful for practical CAV planning in crowded urban environments.
\end{abstract}
\begin{figure*}[htbp]
    \centering
    \includegraphics[width=\linewidth]{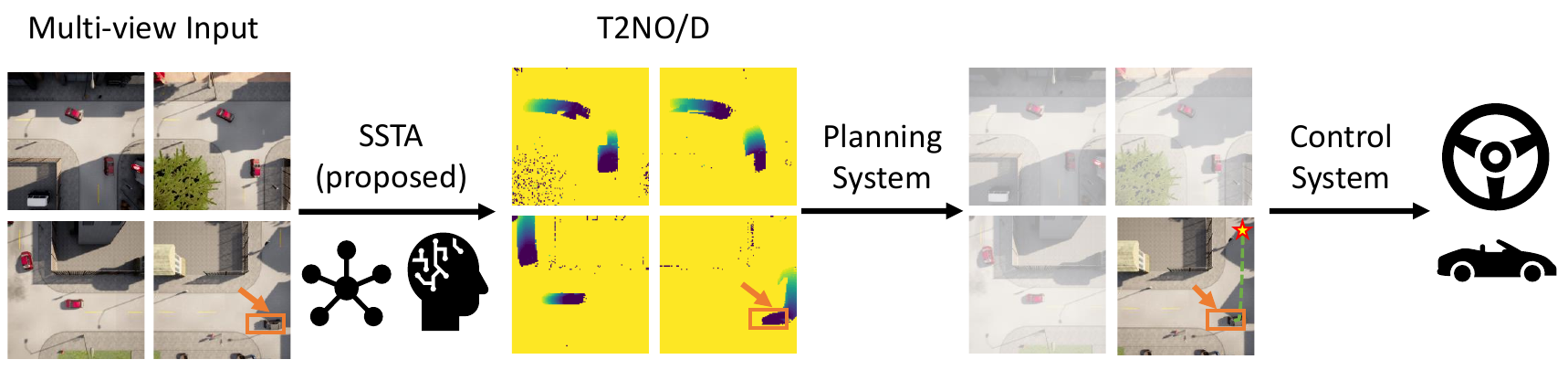}
    \caption{Our proposed system is a distributed Neural Perception System comprised of a network of static traffic cameras paired with their own neural networks. The location of the ego car is marked by the \fcolorbox{orange}{white}{\textcolor{orange}{orange}} box.
    The networks predict future pixel-wise occupancy and departure times in each view, which is our first contribution.
    A planning system then generates a path based on collision checking in the pixel-wise predictions, which is our second contribution.
    Finally, the path plan is handed to a CAV's control system to track.
    }
    \label{fig:framework}
    \vspace{-10pt}
\end{figure*}

\section{Introduction}

Connected and autonomous vehicles (CAVs) have the potential to make urban transportation safer, more efficient, and more sustainable. 
A key challenge towards realizing this goal is that of motion planning; traditional methods rely on static maps and real-time sensor data, and cannot always yield accurate predictions of the future state of the environment due to occlusions or other onboard sensing limitations.
Hence, smart infrastructure has been proposed as a way to augment CAV capabilities.
However, collecting data to train such smart infrastructure can often require prohibitive amounts of manual labeling.
Therefore, there is growing interest in using self-training smart infrastructure to improve CAV safety and performance.

The present work considers a specific type of edge device proposed in our prior work: the Self-Supervised Traffic Advisor (SSTA)~\cite{sun2022self}.
This proposed device automatically processes video data from traffic cameras to support CAV motion planning.
SSTAs are a scalable edge solution that avoid extensive hand-labeling and can provide large-scale connected coverage of an urban environment through self-training to share data.
Our previous work studied the challenges of self-training and networking, whereas this work assesses the potential for SSTAs to benefit CAV motion planning and control.

\textit{Contributions:}
First, we develop a new output format for SSTAs as an alternative to predicting raw video frames.
Specifically, we propose to predict Time to Next Occupancy (T2NO) and Time to Next Departure (T2ND), which describe when each portion of an SSTA's field of view will next be occupied and subsequently unoccupied by a road user.
We abbreviate these together as ``T2NO/D.''
Second, we propose an approach to CAV motion planning that leverages our novel T2NO/D outputs from SSTAs.
Third, we numerically evaluate the effectiveness of autonomously predicting T2NO/D in comparison to raw video within a CAV motion planning framework.
We find that our proposed method can provide an effective smart infrastructure perception output that is directly amenable to CAV planning and control.

\section{Related Work}\label{sec:related_work}

SSTAs~\cite{sun2022self}, and this paper, lie at the intersection of CAV infrastructure, networked learning, and video prediction.

\paragraph{CAV Infrastructure}
To maximize their utility, CAVs should communicate with self-driving and human-driven vehicles, roadside infrastructure, and other road users; many reviews of this literature are available~\cite{rana2021connected,guanetti2018control,pmlr-v155-huang21a,pmlr-v155-sun21a}.
Given bandwidth and connectivity limitations, it is critical to decide what, and when, to communicate.
For example, Gopalswamy et al.~\cite{Swaminathan_8500436} propose to offload CAV responsibilities for situational awareness and information sharing via vehicle-to-infrastructure (V2I) communication.
Alternatively, short-range communications allow vehicle-to-vehicle (V2V) transmission of information such as speed, heading, and brake status~\cite{he2010adaptive,he_jianhua_8888126}.
Since V2V alone can cause network-wide instabilities as the number of vehicles grows \cite{zheng2015stability}, it is critical to explore alternative means of widespread, scalable communication that can improve large-scale traffic metrics such as safety and efficiency \cite{hou2023large}.
In this work, we propose a scalable infrastructure solution and study its potential for aiding CAVs in an isolated V2I manner; we plan to explore mixing V2V and V2I in future work.

\paragraph{Networked Learning}
The concept of networked learning has been recently explored in the context of connected and autonomous vehicles (CAVs) \cite{chellapandi2023survey}.
In this paradigm, multiple CAVs can cooperate to improve their individual driving policies through message passing and sharing of information~\cite{he2020fedml,yu2022dinno,bistritz2020distributed}.
We note that CAVs can cooperate to perceive their surroundings~\cite{yu2022dinno} and to improve traffic congestion and energy efficiency~\cite{9689004,SHI2021103421,9905760}.
In this work, we instead consider how networked learning in \textit{infrastructure} can be used to improve perception and planning for CAVs.

\paragraph{Video Prediction}
Video prediction has the potential to impact traffic management, motion planning for autonomous vehicles, and autonomous surveillance~\cite{walker2016_forecast_var_autoenc_stat_img}.
This task is convenient for data collection because it can often be achieved without manual data annotation.
Deep learning methods can be especially well suited for modeling spatiotemporal relationships in video and generating accurate predictions of future events~\cite{mathieu2015deep,babaeizadeh2017stochastic,vyas2020multi,pan2020cross}. 
However, most previous video prediction is typically focused on a single camera, aiming to learn visual dynamics~\cite{walker2016_forecast_var_autoenc_stat_img} or interpolate video spatially and temporally~\cite{lu2017flexible}.
Our prior work on SSTAs~\cite{sun2022self} instead addressed the challenge of video prediction across a network of \emph{multiple} static traffic cameras.
The key remaining challenge addressed in the present work is that, from a V2I perspective, predicting raw video is not directly amenable to downstream CAV planning and control tasks.
\section{Method}\label{sec:method}

In this section, we describe our system architecture (cf. Fig.~\ref{fig:framework}) and algorithms used for video prediction and motion planning. 
First, we provide
an overview of our proposed method.
Next, we introduce the Time to Next Occupancy (T2NO) and Time to Next Departure (T2ND) to predict future road user trajectories.
We then explain our training loss and neural network architecture used at each SSTA.
Finally, we design a collision-checking technique for T2NO/D and utilize it for downstream planning and control.

\begin{rem}
    In this work, we consider planning only for a single CAV, which we henceforth refer to as the \emph{ego vehicle}.
    We defer multi-agent SSTA-enabled planning for future work.
\end{rem}

\subsection{Overview of Proposed Method}
Our overall goal is CAV motion planning using video predictions generated by SSTAs (Fig.~\ref{fig:framework}).
Motion planning requires collision checking, but performing collision checking directly in raw video predictions is challenging.
Instead, we propose predicting future occupancy in each pixel of each SSTA view. 
Our method has three steps:
(1) calculating future occupancy using Alg.~\ref{alg:compute_T2NO_and_T2ND},
(2) collision checking pixels occupied by the ego vehicle against pixels with predicted future occupancy, and
(3) performing motion planning by modifying the A$^*$ algorithm \cite{Hart1968} to generate an optimal path that avoids intersecting with obstacles in space and time.

\subsection{Time to Next Occupancy and Time to Next Departure}

We leverage the fact that each SSTA is static to capture and use background information.
This enables us to compute two quantities that simplify collision checking: Time to Next Occupancy and Time to Next Departure.

Before defining these quantities, we establish some notation, all within the context of a single SSTA.
Note, to ease notation, we drop the index $i$ for considering the \ith SSTA in this discussion.
Suppose the current time is $t = 0$.
Let $\N_{256} = \{0,1,\cdots,255\}$.
Consider a sequence of predicted images $(I_t)_{t=0}^{T}$ with each image in $\N_{256}^{H\times W\times 3}$ (i.e., height $H$, width $W$, and RGB channel of dimension 3), and $T \in \N$ as the prediction horizon.
Also suppose we also have a \textit{background image} $B \in \N_{256}^{H\times W\times 3}$; in practice we create $B$ for each SSTA by pixel-wise averaging of approximately one minute of video data.
Denote indexing into the $(i,j)$\ts{th} pixel of an image $I$ by $I[i,j]$.
Let $\N_n$ denote the set $\{0,\cdots,n\}$.

\begin{defn}\label{def:T2NO}
    The \emph{Time to Next Occupancy} (T2NO) is the first time at which a given pixel in an SSTA's image is predicted to be occupied, which we assess by comparison against the background $B$.
    We represent T2NO as an array $O \in \R^{H\times W}$.
\end{defn}

\SetKw{Break}{break}

\begin{algorithm}[t]
\caption{Time to Next Occupancy / Departure}
\label{alg:compute_T2NO_and_T2ND}
\KwIn{prediction $(I_t)_{t=0}^T$, background $B$, occupancy threshold $\tau_O$, departure threshold $\tau_D$}

$O, D \gets +\infty_{H\times W}$ // preallocate T2NO and T2ND

\For{$(i,j) \in \N_h\times \N_w$ (i.e., iterate over each pixel)}{
    // first, compute time to next occupancy
    
    \For{$t = 0, 1, \cdots, T$}{
        $\delta \gets |I_t[i,j] - B[i,j]|$
        // error between prediction and background at pixel $(i,j)$
        
        \If{$\delta \geq \tau_O$}{
            $O[i,j] \gets t$
            // get smallest $t$ for which prediction differs from background
            
            \Break
        }
    }

    // second, compute time to next departure
    
    \If{$O[i,j] < +\infty$ (i.e., if there is any future occupancy at the current pixel)}{
        \For{$t = O[i,j], O[i,j]+1, \cdots, T$ (i.e., for all times after the time to next occupancy)}{
            $\delta \gets |I_t[i,j] - B[i,j]|$
            
            \If{$\delta \leq \tau_D$}{
                $D[i,j] \gets t$
                // get smallest $t$ for which prediction is close to background
                
                \Break
            }
        }
    }
}
\KwOut{$O$ (T2NO) and $D$ (T2ND)}
\end{algorithm}

On its own, T2NO does not capture all relevant dynamic information in a scene, because, once a pixel is declared occupied at a time $t$, it is implied to be occupied for all future time $t' > t$, leading to the following definition that alleviates this concern:
\begin{defn}\label{def:T2ND}
    The \emph{Time to Next Departure} (T2ND) measures the first time at which an occupied pixel is no longer occupied.
    Similar to T2NO, we represent T2ND as an array $D \in \R^{H\times W}$.
\end{defn}
We compute both T2NO and T2ND using Alg. \ref{alg:compute_T2NO_and_T2ND}.
By combining T2NO and T2ND, we can assess when each pixel in an SSTA's field of view will be occupied and then free within the prediction horizon $T$.

As an example to clarify these concepts, consider a single car passing through pixel $(i,j)$ at $10~\si{m/s}$, and suppose every pixel corresponds to a $1 \times 1~\si{m^2}$ patch of drivable area.
Suppose the T2NO is $2.0~\si{s}$, indicating the earliest time at which the car occupies that pixel.
Suppose the car is $5~\si{m}$ long.
Then the car will completely pass over the pixel in $0.5~\si{s}$, meaning that the T2ND at that pixel must be no less than $2.5~\si{s}$.
An adjacent pixel in the direction of travel will have $O[i,j] = 2.1~\si{s}$ and $D[i,j] \geq 2.6~\si{s}$ (assuming the car is predicted to maintain its speed and heading).

\subsection{Training to Predict T2NO/D}
We follow the networked co-learning procedure outlined in \cite[Sec. III.D]{sun2022self}.
However, we preprocess each \ith  SSTA's recorded images to generate $O_t^i$ and $D_t^i$ as output by Alg.~\ref{alg:compute_T2NO_and_T2ND} for each time step $t$ of the training data (i.e. past video frames).
Denote the T2NO/D predictions of the \ith SSTA at time ${\hat{t}}$ as $\hat{O}^i_{\hat{t}}$ and $\hat{D}^i_{\hat{t}}$, starting from time $t$ up to a time $t + T$.
We minimize the mean-squared error loss between predicted and ground-truth T2NO/D:
\begin{align}
    \loss_t = \sum_{i=1}^N\left(
            \sum_{{\hat{t}} = t}^{t+T} \left(
                \alpha\norm{O_{\hat{t}}^i - \hat{O}_{\hat{t}}^i}^2 + 
                \beta\norm{D_{\hat{t}}^i - \hat{D}_{\hat{t}}^i}^2
                \right)
        \right),
\end{align}
where $\alpha, \beta > 0$ are tunable hyperparameters, and we use Frobenius norms  (i.e., pixelwise error squared).
Note, the predicted $\hat{O}^i_{\hat{t}}$ and $\hat{D}^i_{\hat{t}}$ are functions of the \ith SSTA's hidden state $h_{\hat{t}}^i$, the set of \textit{received} messages $Y_{\hat{t}}^i$ from its connected SSTAs, and its current received video frame $I_{\hat{t}}^i$.

\subsection{Proposed Neural Network Architecture}
We use the same architecture as in \cite[Sec. III.F]{sun2022self}.
However, instead of requiring a recursive rollout to generate T2NO/D predictions, we generate the prediction with a single forward pass of the network at each SSTA.
So, the outputs of the \ith network at timestep ${\hat{t}}$ are $\hat{O}^i_{\hat{t}}$, $\hat{D}^i_{\hat{t}}$, and the message $y_{\hat{t}}^i$.

\begin{rem}
    The method proposed in this work reduces the number of neural network evaluations at each SSTA from a $T$-step recursive rollout to just a single evaluation, since T2NO/D collapse temporal information into a single image.
\end{rem}

\subsection{Collision Checking with T2NO/D}

We seek to use SSTA predictions for collision checking within a CAV motion planning framework.
First, we explain how to isolate pixels associated with the ego vehicle.
Then, we consider collision checking against T2NO/D.

\begin{table*}[t]
\begin{center}
\renewcommand\tabcolsep{0.5pt}
\begin{tabular}{>{\centering\arraybackslash}m{0.0669\linewidth}|>{\centering\arraybackslash}m{0.0669\linewidth}?>{\centering\arraybackslash}m{0.0669\linewidth}|>{\centering\arraybackslash}m{0.0669\linewidth}?>{\centering\arraybackslash}m{0.0669\linewidth}|>{\centering\arraybackslash}m{0.0669\linewidth}?>{\centering\arraybackslash}m{0.0669\linewidth}|>{\centering\arraybackslash}m{0.0669\linewidth}?>{\centering\arraybackslash}m{0.0669\linewidth}|>{\centering\arraybackslash}m{0.0669\linewidth}?>{\centering\arraybackslash}m{0.0669\linewidth}|>{\centering\arraybackslash}m{0.0669\linewidth}?>{\centering\arraybackslash}m{0.0669\linewidth}|>{\centering\arraybackslash}m{0.0669\linewidth}}
\toprule
 \multicolumn{2}{c?}{Ref Img $B$}  &  \multicolumn{2}{c?}{Cur Image $I$}  &  \multicolumn{2}{c?}{$T=2$}  & \multicolumn{2}{c?}{$T=10$}  & \multicolumn{2}{c?}{$T=20$} & \multicolumn{2}{c?}{$T=30$}  & \multicolumn{2}{c}{$T=60$}   \\
\midrule
 \includegraphics[width=\linewidth]{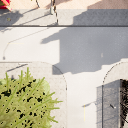} &  \includegraphics[width=\linewidth]{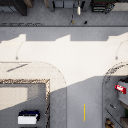} &
  \includegraphics[width=\linewidth]{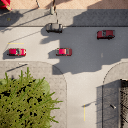} &  \includegraphics[width=\linewidth]{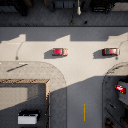} &
 \includegraphics[width=\linewidth]{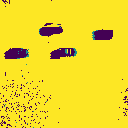} &  \includegraphics[width=\linewidth]{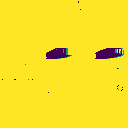} &
 \includegraphics[width=\linewidth]{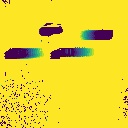} &  \includegraphics[width=\linewidth]{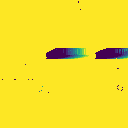} & 
 \includegraphics[width=\linewidth]{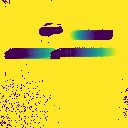} & \includegraphics[width=\linewidth]{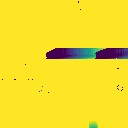} & \includegraphics[width=\linewidth]{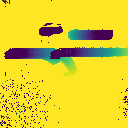} & \includegraphics[width=\linewidth]{fig/time_horizon/vis_t2no__out_3_30_00000525_plt.png}  & \includegraphics[width=\linewidth]{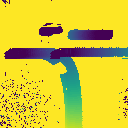} & 
\includegraphics[width=\linewidth]{fig/time_horizon/vis_t2no__out_3_60_00000525_plt.png}
\\ 
\midrule
\includegraphics[width=\linewidth]{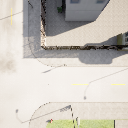} &  \includegraphics[width=\linewidth]{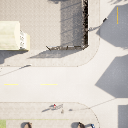} &
\includegraphics[width=\linewidth]{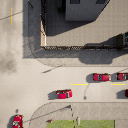} &  \includegraphics[width=\linewidth]{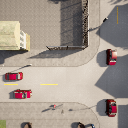} &
\includegraphics[width=\linewidth]{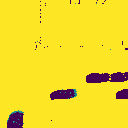} &  \includegraphics[width=\linewidth]{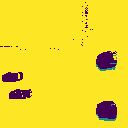} &
\includegraphics[width=\linewidth]{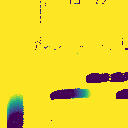} &  \includegraphics[width=\linewidth]{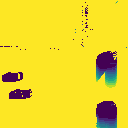} & 
\includegraphics[width=\linewidth]{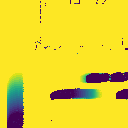} & \includegraphics[width=\linewidth]{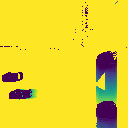} & \includegraphics[width=\linewidth]{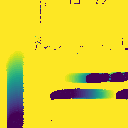} & \includegraphics[width=\linewidth]{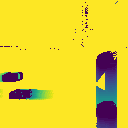}  & \includegraphics[width=\linewidth]{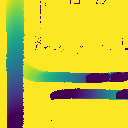} & 
\includegraphics[width=\linewidth]{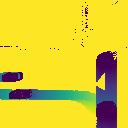} 
\\
\bottomrule
\end{tabular}
\captionof{figure}{Visualization of T2NO for different time horizons (T2ND is omitted for ease of understanding).
The color scale runs from black at $t = 0$ to yellow as $t \to \infty$.
The background image $B$ and the current image $I$ are also shown on the left.
We see that T2NO contains little time information when the time horizon is short, whereas T2NO contains ambiguous information when the time horizon is long, and different trajectories even overlap, impeding the effectiveness of CAV motion planning. Even though there are multiple plausible paths for cars to take in the future, depending on the road topology, this ambiguity is not desired and hinders planning.}
\label{fig:pred_vis}
\end{center}
\vspace{-15pt}
\end{table*}

\subsubsection{Masking the Ego Vehicle's Pixels}

Recall that each SSTA outputs a sequence of images $(I_t)_{t=0}^T$, with each $I_t \in \N_{256}^{h\times w \times 3}$.
We now introduce an approach to extracting the pixels in an image corresponding to our ego vehicle's pose (2-D position and 1-D heading), which we denote $(p_t,\theta_t)$ at time $t$.
Suppose the vehicle has a rectangular body of dimensions $l \times d$.
Also suppose that each pixel, with indices $(i,j)$, is associated with a corresponding spatial location $p_{i,j}$, which can be found with the following map:
\begin{align}
    \code{sp2px}&: p_{i,j} \mapsto (i,j)
\end{align}
Since each SSTA is a static camera with a known field of view, it is straightforward to create this map via standard camera calibration techniques.
Also, note that we can represent the vehicle's body at the state $(p,\theta)$ as an array of vertices arranged counterclockwise:
\begin{align}
    V_t \gets \left(\tfrac{1}{2}
        \left[\begin{smallmatrix*}[r]
            \cos\theta_t & -\sin\theta_t\\
            \sin\theta_t & \cos\theta_t
        \end{smallmatrix*}\right]
        \left[\begin{smallmatrix*}[r]
            l & -l & -l & l \\
            d & d & -d & -d
        \end{smallmatrix*}\right]\right) +
        p_t
\end{align}
Recall that our images $I_t$ are of dimension $h \times w \times 3$.
We thus create a mask for the given vehicle pose:
\begin{align}
    P_t &\gets \code{sp2px}(V_t) \\
    \label{eq:fill_poly}
    P_t &\gets \code{fillPoly}(\zeros_{h \times w \times 3},\ P_t), \\
    M_t &\gets P_t == (255,255,255),\label{eq:vehicle_mask}
\end{align}
where $P_t$ is used as an intermediate array.
Equation~\eqref{eq:fill_poly} fills in a polygon defined by the (pixel coordinate) vertices $V_M$, into a black image ($\zeros_{h \times w \times 3}$), so $P_t \in \N_{256}^{H\times W\times 3}$.
Note that \code{fillPoly} is available in OpenCV~\cite{opencv_library}, and we assume the default fill color is white, or $(255,255,255)$.
Equation~\eqref{eq:vehicle_mask} creates the mask $M_t \in \{0,1\}^{H\times W}$ (i.e., a Boolean array) which is \code{true} for every entry (i.e., pixel) that is white, thereby identifying the pixels corresponding to our vehicle.

\subsubsection{Collision Checking Against T2NO/D}
Suppose we have the T2NO/D arrays  $O \in \R^{H\times W}$ and $D \in \R^{h \times w}$ output from Alg.~\ref{alg:compute_T2NO_and_T2ND}.
For each time step $t$, we check:
\begin{align}
    \label{eq:convert_mask}
    C_t &\gets t \cdot M_t \\
    \code{chk}_t & \gets \code{any}\left(C_t \geq O\ \&\ M_t \leq D \right)\label{eq:collision_check}
\end{align}
Equation~\eqref{eq:convert_mask} converts the Boolean mask $M_t$ to a real-valued array, where entries for which $M_t$ is \code{true} are set to $t$, and all others are $0$.
Then, we directly check the entries of $M_t \in \R^{H\times W}$ against $O$ and $D$.
If any of the vehicle's occupied pixels (i.e., \code{true} pixels) overlap with the time window created by $O$ and $D$, then $\code{chk}_t$ is \code{true} (i.e., the pose $(p_t,\theta_t)$ is in a collision).

\subsection{Planning with T2NO/D}\label{subsec:planning_with_T2NO}
After obtaining T2NO/D, we plan a path for our ego vehicle.
In this work, we extend the classical A$^*$ algorithm \cite{Hart1968} to leverage T2NO/D, but we note that these representations can be readily applied to other planning algorithms.

To avoid collisions with dynamic obstacles, we modify A$^*$ to generate a time-optimal path that avoids intersecting with the obstacles in both time and space.
Suppose the ego vehicle is at a time and position $z_0 = (0,p_0) \in [0,T]\times\R^2$.
Our implementation finds a list of times and positions $z_i = (t_i,p_i)$, where $i \in \mathbb{N}$, which define a trajectory for the ego vehicle to track.
We seek to reach a user-specified goal position $p\goal$ in the shortest time while avoiding other vehicles.
As per the classical A$^*$ notation, we use an evaluation function to represent running cost:
\begin{align}
    f(z_i) = c(z_i) + r_t(z_i)
\end{align}
where $c(z_i)$ is the cost of the path from $z_0$ to $z_i$ and $r_t$ is a time-varying heuristic function that estimates the cost of the most efficient path from $p_i$ to $p\goal$ that avoids dynamic obstacles.
In particular, we set
\begin{align*}
    r_t(z_i) = \norm{p_i - p\goal} + K\cdot\code{chk}_{t_i},
\end{align*}
where $K > 0$ is a scalar that penalizes being in collision at time $t_i$, and $\code{chk}_{t_i}$ is the collision check evaluated at time $t_i$ as per \eqref{eq:collision_check} (and $\code{true} = 1$).
The final output of A$^*$ is a discrete sequence of times and positions.
We perform cubic spline interpolation to smooth the sequence before handing it to the ego vehicle to track (in our implementation, the ego vehicle uses a default PD controller from the CARLA simulator \cite{dosovitskiy2017carla}).

Next, we evaluate our proposed technical approach via numerical experiments.

\section{Experiments}\label{sec:experiments}

We now assess the utility of SSTAs and our proposed T2NO/D output format via a series of numerical experiments.
We find that SSTAs can generate predictions which are effective for CAV motion planning.

\begin{figure}[]
    \centering
    \subfloat[Histogram of Ego Vehicle Speed]{\includegraphics[width=0.49\linewidth]{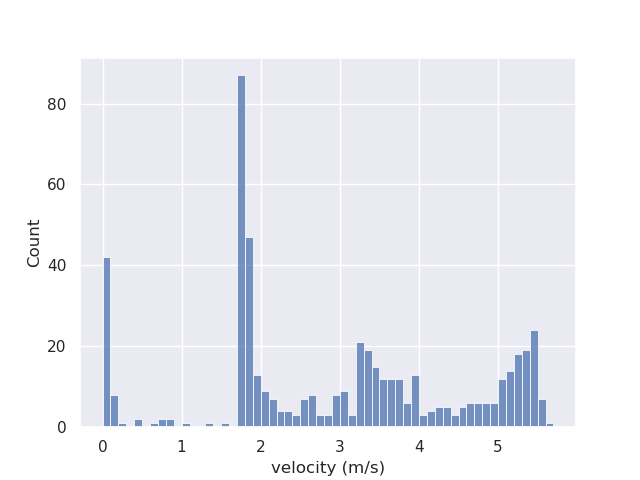}\label{fig:hist_a}}
    \subfloat[Speed Over Time]{\includegraphics[width=0.49\linewidth]{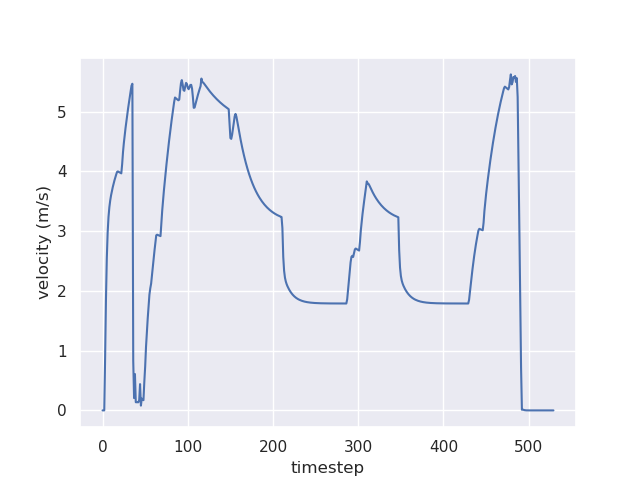}\label{fig:hist_b}}
    \caption{\textbf{Ego Car Speed Statistics}
    }
    \label{fig:hist_of_speed}
    \vspace{-10pt}
\end{figure}

\begin{table*}[htbp]
\caption{Performance in 4-view setting over 20 trials (mean $\pm$ standard deviation).}
\begin{center}
\begin{tabular}{>{\centering\arraybackslash}m{0.1\linewidth}|>{\centering\arraybackslash}m{0.11\linewidth}|>{\centering\arraybackslash}m{0.11\linewidth}|>{\centering\arraybackslash}m{0.12\linewidth}|>{\centering\arraybackslash}m{0.11\linewidth}|>{\centering\arraybackslash}m{0.11\linewidth}|>{\centering\arraybackslash}m{0.11\linewidth}}
\hline
\textbf{Metrics} & Success Rate (\%) $\uparrow$ & Number of Collisions $\downarrow$ & Total Timesteps to Reach Goals $\downarrow$ & Total Control Effort $\downarrow$ & Total Sudden Reversals $\downarrow$ & Travel Distance (km) $\uparrow$ \\
\hline
\textsc{Monolith} & $\bm{65.0}$ & $15.15 \pm \bm{0.86}$ & $\bm{484.15}\pm 6.72$ & $\bm{245.49} \pm 5.91$ & $66.76 \pm 1.36$ & $\bm{1.21 \pm 0.34}$  \\
\textsc{AllComms} & 60.0 & $\bm{15.00} \pm 4.63$  & $495.50\pm 6.76$ & $251.47 \pm \bm{4.01}$ & $\bm{66.58\pm 1.25}$  & $1.12 \pm 0.35$  \\
\textsc{NoComms} & 50.0 & $17.50 \pm 2.29$ & $508.40 \pm \bm{5.58}$ & $260.28 \pm 4.33$ & $70.50 \pm 2.94$ & $1.04 \pm 0.38$ \\
\hline
\end{tabular}
\label{tab:4-view-performance}
\end{center}
\vspace{-10pt}
\end{table*}

\subsection{Experiment Setup}

\subsubsection{Implementation Details}
We evaluate our approach on a desktop computer with an NVIDIA GeForce RTX 2080Ti and an Intel Core i7-6800K CPU.
We design the traffic scenario and evaluate our pipeline with the CARLA 0.9.11 simulator.
We train our networks using the ADAM optimizer~\cite{DBLP:journals/corr/KingmaB14} and employ a mini-batch size of 10 sequences.
The learning rate is set to $10^{-3}$, and training is stopped after 800 epochs.

\subsubsection{SSTAs}
In each simulated traffic scenario, SSTAs are placed at each intersection and record top-down images. 
Each SSTA communicates with its neighbors at adjacent intersections, according to the topology of the road graph. 
Following prior work~\cite{sun2022self}, SSTAs are trained to predict traffic flow (see Fig.~\ref{fig:pred_vis}) using a corpus of simulated traffic scenarios offline. Note that in contrast our prior work explored the effects of online and lifelong learning.

\subsubsection{Ego Vehicle}
We perform planning for an ego vehicle as per Sec. \ref{subsec:planning_with_T2NO}.
The overall task is to avoid other vehicles while attempting to reach sub-goals or a global goal location.
To ease identification of the ego vehicle,
we set its color to grey, and all other vehicles to red for visualization. 
The outcomes are presented in Tab.~\ref{tab:4-view-performance}. 
Since T2NO/D is computed based on background subtraction, the actual vehicle color does not impact our proposed method's performance, which is an improvement over our prior work.

The ego vehicle is initialized with the same start and goal locations in each experiment, but the traffic pattern is randomly generated each time.
An experiment is terminated once the first vehicle reaches its goal location, and we record this time $\tfirst$ as well as the corresponding distance between each other vehicle and its goal as $\davg$.

The ego vehicle utilizes a replanning frequency of $20~\si{Hz}$ (i.e., it plans in a receding-horizon manner). This necessitates the SSTAs to update their prediction outputs at the same frequency. The maximum speed of the ego vehicle is set to the CARLA default of 8.33 m/s. Please refer to Fig.~\ref{fig:hist_of_speed} for a histogram plot displaying the actual speeds achieved by the vehicle in an experiment.

\subsubsection{Metrics}

We evaluate our approach on safety, efficiency, and smoothness with the following metrics:
\begin{itemize}
    \item \textit{Success Rate} is the percentage of experiments in which the ego car is able to reach the global goal position.
    
    \item \textit{Number of Collisions} is how many times the ego car collides with other vehicles or obstacles before reaching its global goal. Note that in our experiments, successful navigation may entail encountering multiple non-fatal collisions to simulate a realistic and challenging environment, consistent with the default setting in CARLA~\cite{dosovitskiy2017carla}. 
    This approach aims to emulate real-world conditions where occasional collisions may happen without compromising the overall navigation capabilities of the autonomous vehicle.

    \item \textit{Travel Distance} is the total distance that the ego car is able to travel before experiencing a collision or reaching its goal.
    
    \item \textit{Total Timesteps to Reach Goal} measures the duration taken to reach the goal, if it was reached.
    
    \item \textit{Total Control Effort} is the sum of the norm of the ego vehicle's 2-D acceleration over the entire experiment.
    That is, if the vehicle experiences an acceleration $\ddot{x}_t$ at time $t$, then this metric is $\sum_{t=0}^{t\lbl{final}} \|\ddot{x}_t\|$, where $t\lbl{final}$ is the final time in the given experiment (i.e., when the vehicle reached the goal or crashed)
    
    \item \textit{Total Sudden Reversals} counts how often the vehicle's longitudinal and lateral acceleration changes sign over the course of an experiment.
\end{itemize}
\looseness=-1
Note, we only count successful trials (i.e., goal reached) for all metrics except for Success Rate and Travel Distance.
This is because failed trials result in smaller Total Control Effort and Total Sudden Reversals due to shorter distances, but this does not mean that failed trials are better than successes.
Also, there is no Total Timesteps to Reach Goal for unsuccessful trials.
\subsection{Independent Variables and Hypotheses}
\subsubsection{Influence of Communication}
We aim to characterize the effects of SSTA communication upon traffic flow using the metrics above.
As such, we compare the performance of our method using the following communication arrangements:
\begin{itemize}
    \item \textsc{Monolith}, in which a single model is trained on all images concatenated together, with no need to communicate, as opposed to multiple models trained at each location and requiring communication.
    \item \textsc{AllComms}, in which each SSTA can communicate with \emph{all} others both during training and at run time.
    \item \textsc{NoComms}, in which SSTAs are trained completely independently and cannot communicate with one another either at training time or run time.
\end{itemize}
Note, as a control on the experiment, we also evaluated a \textsc{NoReplan} baseline, in which agents only receive the first SSTA prediction and hence never replan.
This baseline was never able to reach the goal, as expected, and is therefore not included in the results below.
\textit{Hypothesis:} performance will increase as communication increases, with the \textsc{Monolith} framework performing best.

\subsection{Results and Discussion}

\subsubsection{Influence of Communication}
From the results shown in Tab.~\ref{tab:4-view-performance}, our framework can effectively guide the vehicles to reach the goal for 4-view scenarios.
The \textsc{NoComms} and \textsc{Monolith} baselines provide lower and upper limits for SSTA performance, respectively.
We see that, by communicating with adjacent units, SSTAs' predictions still capture most of the useful patterns in traffic flow; the performance of our approach is only slightly worse than that of \textsc{Monolith}, and substantially superior to that of \textsc{NoComms}.

\subsubsection{Failure Cases}
\begin{figure}[]
    \centering
    \includegraphics[width=\linewidth]{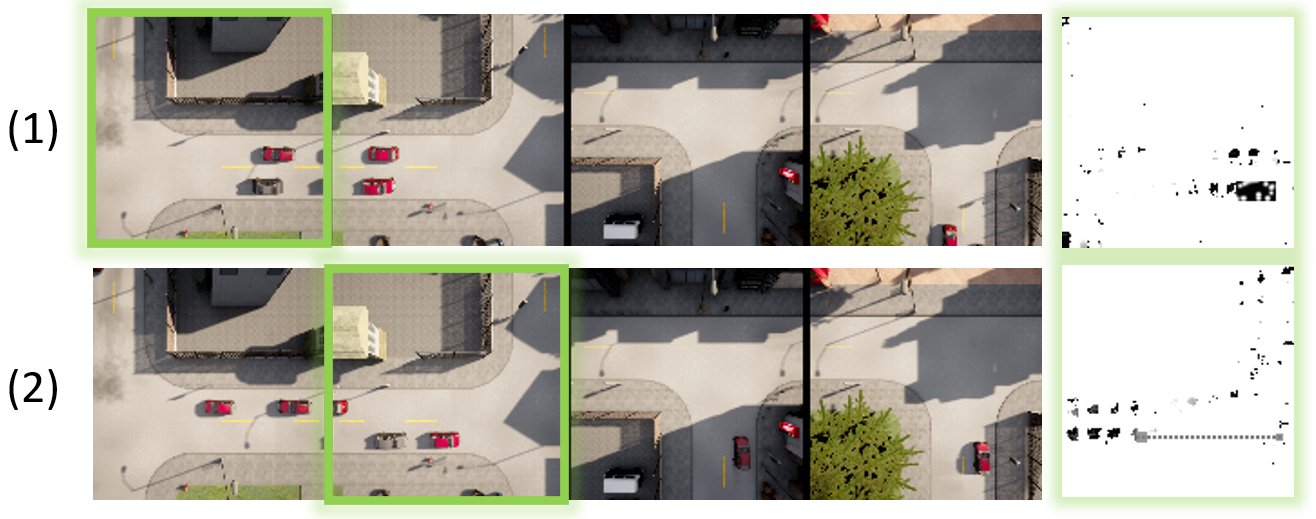}
    \caption{\textbf{Failure Cases}: two failure modes: (1) misjudging the distance to the vehicle ahead of the ego car, resulting in an unnecessary braking maneuver, and (2) failing to predict the presence of an obstacle and causing a collision.}
    \label{fig:failure_case}
    \vspace{-10pt}
\end{figure}

In Fig.~\ref{fig:failure_case}, we show some examples of the predicted results and corresponding planned path using our pipeline.
We find that, for \textsc{NoComms}, the prediction is not accurate enough, which causes two failure modes:
(1) misjudging the distance to the vehicle ahead of the ego car, resulting in an unnecessary braking maneuver, and
(2) failing to predict the presence of an obstacle and causing a collision.

\subsubsection{Practical Considerations}
In addition, to assess the conservativeness of our method, we analyze the vehicle speeds for all trajectories in a histogram and a velocity profile, both shown in Fig.~\ref{fig:hist_of_speed}.
The vehicle spends a substantial time traveling at a variety of speeds, as expected in dense urban traffic, and is not excessively conservative.

\subsubsection{Main Takeaways}
Overall, our experiments confirm the utility of SSTAs for aiding in CAV motion planning.
We note that these experiments relied entirely on the SSTA off-board sensor for ego-motion planning.
Hence, we see much more collisions than would be expected for a safe, practical CAV.
In a real-world deployment, we would prescribe SSTA outputs as an \textit{augmentation} of CAV sensor data; to isolate the benefits of SSTAs, we ignore all onboard ego vehicle sensors and computation for our experiments.

\section{Conclusion}\label{sec:conclusion}

This paper further develops SSTAs \cite{sun2022self} by introducing a learning framework for self-supervised traffic prediction paired with planning vehicle motion in a smart city.
Our experimental results indicate that the proposed framework makes progress towards three goals: (1) by communicating with one another, SSTAs can make more accurate forecasts of traffic flow, (2) CAVs can use SSTA predictions outputs for downstream motion planning and control, and (3) this pipeline can improve traffic metrics such as efficiency and safety.
Of course, limitations remain; future work will improve prediction quality, especially in the transition region between different views, seek to detect anomalous data (i.e., unpredictable traffic events), and explore notions of road user privacy.
Our work represents a step forward in the development of CAV motion planning algorithms in partnership with smart infrastructure, with the long-term goal of creating safer, more efficient, and more sustainable urban transportation systems.

\section*{Acknowledgment}
Toyota Research Institute provided funds to support this work.

\renewcommand{\bibfont}{\normalfont\footnotesize}
{\renewcommand{\markboth}[2]{}
\printbibliography}

\end{document}